\documentclass[10pt, a4paper]{article}
\usepackage{lrec2022} 
\usepackage{multibib}
\newcites{languageresource}{Language Resources}
\usepackage{graphicx}
\usepackage{tabularx}
\usepackage{soul}
\usepackage{titlesec}
\titleformat{\section}{\normalfont\large\bfseries\center}{\thesection.}{1em}{}
\titleformat{\subsection}{\normalfont\SmallTitleFont\bfseries\raggedright}{\thesubsection.}{1em}{}
\titleformat{\subsubsection}{\normalfont\normalsize\bfseries\raggedright}{\thesubsubsection.}{1em}{}
\renewcommand\thesection{\arabic{section}}
\renewcommand\thesubsection{\thesection.\arabic{subsection}}
\renewcommand\thesubsubsection{\thesubsection.\arabic{subsubsection}}

\usepackage{epstopdf}
\usepackage[utf8]{inputenc}

\usepackage{hyperref}
\usepackage{xstring}

\usepackage{color}

\usepackage{xspace}
\usepackage{fancyvrb} 
\usepackage{enumitem} 		   
\usepackage{tabularx,booktabs} 
\usepackage{bm}				   
\usepackage{mathtools} 		   
\usepackage{color,soul}        
\usepackage{amsmath}           
\usepackage{tikz}              
\usepackage{esvect}  
\usepackage{pifont}
\usepackage{adjustbox,fancyvrb,framed} 
\usepackage{tikz}              
\usepackage{mathtools} 		   
\usepackage{amsfonts} 

\makeatletter
\DeclareRobustCommand*\cal{\@fontswitch\relax\mathcal}
\makeatother

\newcolumntype{L}{>{\raggedright\let\newline\\\arraybackslash\hspace{0pt}}m{0.45\linewidth}}

\title{VaccineLies: A Natural Language Resource for Learning to Recognize Misinformation about the COVID-19 and HPV Vaccines}

\name{Maxwell~A.~Weinzierl, Sanda~M.~Harabagiu} 

\address{Human Language Technology Research Institute, Department of Computer Science, The University of Texas at Dallas \\
         800 W. Campbell Rd., Richardson, TX 75080 \\
         \{maxwell.weinzierl, sanda\}@utdallas.edu\\}

\abstract{
Billions of COVID-19 vaccines have been administered, but
many remain hesitant. Misinformation about the COVID-19 vaccines and other vaccines, propagating on social media, is believed to drive hesitancy
towards vaccination. The ability to automatically recognize misinformation targeting vaccines on Twitter depends on the availability of data resources. In this paper we 
present {\sc VaccineLies}, a large collection of tweets propagating misinformation about two vaccines: the COVID-19 vaccines and the Human Papillomavirus (HPV) vaccines. Misinformation targets are organized in vaccine-specific taxonomies, which reveal the misinformation themes and concerns. The ontological commitments of the Misinformation taxonomies provide an understanding of which misinformation themes and concerns dominate the discourse about the two vaccines covered in {\sc VaccineLies}. The organization into training, testing and development sets of {\sc VaccineLies} invites the development of novel supervised methods for detecting misinformation on Twitter and identifying the stance towards it. Furthermore, {\sc VaccineLies} can be a stepping stone for the development of datasets focusing on misinformation targeting additional vaccines.\\ \newline  }

\begin{document}

\maketitleabstract

\section{Introduction}

Misinformation spreading, especially on social media, is believed to be responsible for vaccine hesitancy \cite{covid-misinfo-epidemic}. It is imperative for public health practitioners to know {\em what} misinformation is spreading as well as who is {\em adopting} or {\em rejecting} it, such that interventions can be tailored appropriately. 
Public health messaging approaches could not only inoculate against misinformation, but also effectively reach social media users with promise to shift or bolster vaccine attitudes.
However, developing natural language processing methods targeting the identification of misinformation about vaccines in social media postings suffers from the lack of language resources where vaccine misinformation annotations are available. 

When misinformation was cast as rumor detection, several well-known benchmark datasets for misinformation detection on Twitter became available. For example, the Twitter15 \cite{Ma-ijcai16}
and Twitter16 \cite{ma-etal-2017-detect} datasets consist of a collection of tweets annotated as true rumors, false rumors, unverified rumors or non-rumors. Unfortunately, they did not cover any vaccine misinformation. Similarly,
the PHEME dataset \cite{Zubiaga2016} consists of
Twitter conversation threads 
making a true and false claim, and a series of replies, but none of the conversations focused on vaccination. {\sc COVIDLies} \cite{covidlies} is a dataset
generated from 86 known misconceptions about COVID-19, for which tweets that evoke the misconceptions were retrieved and annotated with their {\em stance} towards the misconceptions. Inspired by {\sc COVIDLies}, we have created  {\sc VaccineLies}, a dataset which addresses misinformation about two different vaccines: the COVID-19 vaccines and the vaccines protecting against the Human Papillomavirus (HPV).

\begin{table}[h]
\centering
\small
\begin{tabular}{p{0.45\textwidth}}
    \toprule 
    {\bf Misinformation Target:} {\em COVID-19 vaccine alters DNA. }\\
    \midrule
    {\sc Stance:} {\bf Accept}\\
    {\em Tweet:} 
    @USER Good girl. The COVID-19 vaccination is an mRNA vaccine, this means it alters your body's DNA. This is incredibly dangerous, as no one knows the long term effects of this.
    \\
    \hline
    {\sc Stance:} {\bf Reject} \\
    {\em Tweet:} 
    @USER This is absolutely false. The mRNA from a COVID-19 vaccine never enters the nucleus of the cell, which is where our DNA is kept. The mRNA does not affect or interact with our DNA in any way.  If you have other concerns, then fine, but don’t fall for unfounded nonsense.
    \\
    \midrule
    {\bf Misinformation Target:} {\em HPV vaccine was banned. }\\
    \midrule
    {\sc Stance:} {\bf Accept}\\
    {\em Tweet:} 
    @USER Don't worry most people I know don't take the vaccination. We've seen the countries that have banned the hpv vaccine because Gates drug was maiming and killing young girls!
    \\
    \hline
    {\sc Stance:} {\bf Reject} \\
    {\em Tweet:} 
    @USER Excuse me, actual person from Europe here you interfering trollop. WHAT vaccines are banned in the EU? A UK company released its own HPV vaccine that got outcompeted by Gardasil as it targeted a larger array of strains. Don’t involve us to push your primitive agenda
    \\ 
    \bottomrule
\end{tabular}
\caption{ Misinformation Targets for the COVID-19 and HPV vaccines with tweets evoking them.}
\label{tb:examples}
\end{table}
\raggedbottom

We present {\sc VaccineLies}, which consists of: \\
$\lhd$1$\rhd$ {\em Misinformation Targets} (MisTs) similar to those illustrated in Table~\ref{tb:examples}, addressing misinformation towards COVID-19 or HPV vaccines;\\ 
$\lhd$2$\rhd$ the {\em tweet IDs} for those tweets that were judged as evoking any of the MisTs available in {\sc VaccineLies};  \\
$\lhd$3$\rhd$ annotation of the {\em stance} of each tweet author that evoked a MisT, indicating if they {\em Accept} the Mist; {\em Reject} it, or they have no stance towards it. \\
$\lhd$4$\rhd$ a {\em taxonomy} of the MisTs,  which enables the interpretation of the themes and concerns characterizing the vaccine misinformation available in {\sc VaccineLies}. The taxonomical organization into themes and concerns of the misinformation targets for each vaccine 
will illuminate the discovery of which targets of misinformation dominate when the vaccines are discussed in social media and, in addition, will lead to the discovery of which kinds of vaccine misinformation are most adopted or most rejected in {\sc VaccineLies}. 
Separate misinformation taxonomies were discerned for the COVID-19 vaccine and the HPV vaccine.
{\sc VaccineLies} was inspired by {\sc COVIDLies} \cite{covidlies}, a dataset of Twitter annotations focusing on misinformation about COVID-19. Like in {\sc COVIDLies}, we use the notion of Misinformation Target (MisT) to refer to misconceptions that are employed for propagating misinformation. In addition to misconceptions, we considered misinformation
any reference to conspiracy theories or any flawed reasoning. Moreover, we extended the
methodology for identifying MisTs, relying not only on misinformation that is readily available on Wikipedia web pages, but also on misinformation that is widely discussed in Twitter conversations. In addition to providing a set of MisTs focusing on two different vaccines, {\sc VaccineLies} provides a
large set of IDs for tweets that evoke at least one of the MisTs, which were judged by language experts as being relevant to the misinformation expressed in MisTs. Furthermore, the stance
of the author of each tweet that evokes a MisT was judged, indicating whether the author {\em Accepts} the MisT, because they agree with it; {\em Rejects} the MisT, as they disagree with it; or the author has {\em No Stance} towards the MisT, although it is evoked.

The annotations that enabled the creation of {\sc VaccineLies} were performed jointly by language experts from The University of Texas at Dallas and public health experts from The University of California, Irvine.
In previous work \cite{weinzierl-covid-glp}, we used an earlier version of {\sc VaccineLies} to perform automatic detection of COVID-19 misinformation on Twitter. We also used {\sc VaccineLies} to identify vaccine hesitancy profiles of users on Twitter \cite{weinzierl-scaling-discovery}. To our knowledge, {\sc VaccineLies} is the only publicly available resource tackling misinformation about the COVID-19 and HPV vaccines on Twitter.
We believe our annotation efforts in constructing {\sc VaccineLies} fills a gap in vaccine misinformation research, which could greatly benefit both public health experts and natural language processing researchers.

{\sc VaccineLies} can be also seen as consisting of two vaccine-specific datasets, 
namely {\sc CoVaxLies} and {\sc HpVaxLies}, 
corresponding to their focus on misinformation concerning the COVID-19 or the HPV vaccine, respectively. This organization of {\sc VaccineLies} presents the advantage that it allows language researchers and public health experts to contemplate efforts of bootstrapping the discovery of misinformation targeting other vaccines on social media. 
The remainder of the paper is organized as follows. Section~\ref{sec:mists} introduces the  process for identifying Misinformation Targets (MisTs), while Section~\ref{sec:tax} details the organization of MisTs into misinformation taxonomies targeting the COVID-19 and HPV vaccines.
Section~\ref{sec:collection} describes the methodology used for recognizing tweets which evoke any of the MisTs, while 
Section~\ref{sec:annotation} presents the {\em stance} annotation process used in {\sc VaccineLies}. 
Section~\ref{sec:methods} describes a cross-vaccine transfer learning approach and Section~\ref{sec:results} presents and discusses the  experimental results for cross-vaccine transfer learning. 
Section~\ref{sec:conclusion} summarizes the conclusions.

\section{Vaccine Misinformation Targets}
\label{sec:mists}

\begin{figure*}[ht]
    \centering
    \includegraphics[width=0.8\textwidth]{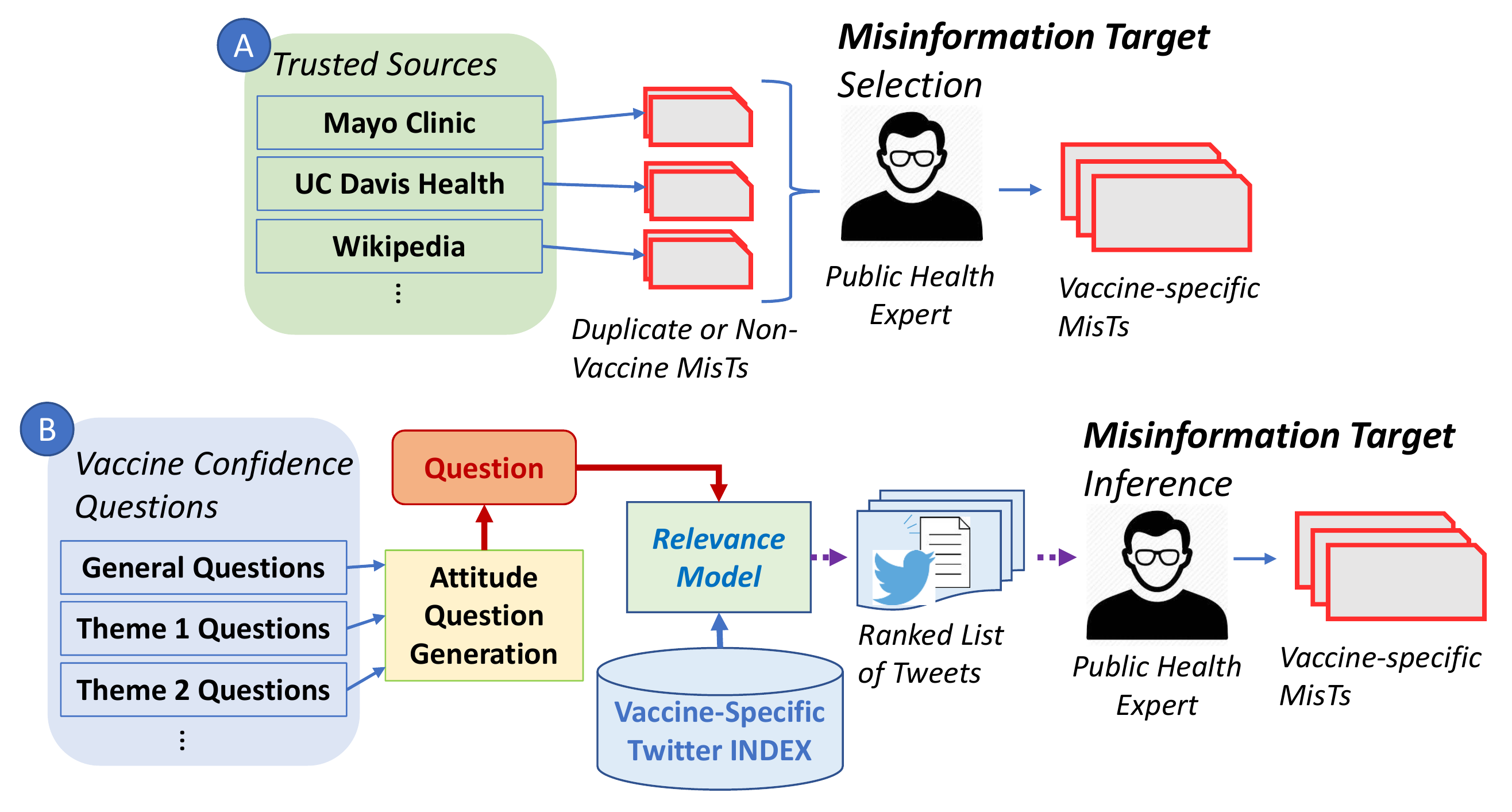}
    \caption{Misinformation Target (MisT) discovery utilizing (A): Trusted sources of vaccine misinformation, and (B): Our Question/Answering framework for vaccine misinformation discovery.}
    \label{fig:mists}
\vspace{-4mm}
\end{figure*}

In {\sc VaccineLies} the identification of misinformation targeting the COVID-19 and HPV vaccines on Twitter was performed in two different ways. First, we have considered several 
trusted sources, such as the Mayo Clinic, University of California (UC) Davis Health, as well as the Wikipedia page available at {\em en.wikipedia.org/wiki/COVID-19\_misinformation\#Vaccines}, as illustrated in Figure~\ref{fig:mists} (A). 
These trusted sources have been actively collecting and debunking misinformation about COVID-19 since the beginning of the pandemic, and much of this misinformation is about the COVID-19 vaccine.
MisTs from these trusted sources were merged into a final collection of 17 MisTs targeting the COVID-19 vaccines.
However, the identification of trusted sources that debunk misinformation was more challenging for the HPV vaccine. For example, there is no Wikipedia page dedicated to misinformation about the HPV vaccine, although there is a page that is dedicated to the vaccine. There is a Wikipedia page dedicated to vaccine misinformation in general listing several misinformation themes, however, it did not provide specific MisTs for the HPV vaccine.

A second approach, illustrated in Figure~\ref{fig:mists}(B), was considered, which utilized questions from the Vaccine Confidence Repository \cite{Rossen} to find answers from a vaccine-specific index of unique tweets obtained from the Twitter API. Whenever these answers  contain misinformation about either the COVID-19 or HPV vaccines, they
were considered MisTs. 
Before using the second approach,
approval from the Institutional Review Board at the University of Texas at Dallas was obtained in order to use the Twitter API to collect tweets discussing either the HPV or the COVID-19 vaccine: IRB-21-515 stipulated that our research met the criteria for exemption \#8(iii) of the Chapter 45 of Federal Regulations Part 46.101.(b). 

Tweets discussing either the COVID-19 or the HPV vaccines were obtained by querying the Twitter API.
A collection of 9,133,471 tweets was obtained from the Twitter streaming API as a result of the query {\em ``(covid OR coronavirus) vaccine lang:en -is:retweet”}. We perform Locality Sensitive Hashing (LSH) \cite{lsh} with term trigrams, 100 permutations, and a Jaccard threshold of 50\%, to produce 5,865,046 unique tweets discussing COVID-19 vaccines. 
These tweets were authored in the time frame from December 18th, 2019, to July 21st, 2021.
Similarly, a collection of 864,008 tweets was obtained from the Twitter historical API as a result of the query {\em 
``(human papillomavirus vaccination) OR (human papillomavirus vaccine) OR gardasil OR cervarix OR (hpv vaccine) OR (hpv vaccination) OR (cervical vaccine) OR (cervical vaccination) lang:en -is:retweet"}. After using LSH for detecting near-duplication, we
obtained  422,078 unique tweets discussing HPV vaccines.
Both tweet collections were organized in vaccine-specific indexes, obtained  using Lucene \cite{lucene} with the BM25 vector relevance model \cite{bm25}, which informed the Q/A framework illustrated in  Figure~\ref{fig:mists}(B).

The questions that were asked originate in the Vaccine Confidence Repository (VCR) \cite{Rossen}. For each of the 19 questions available in VCR, we generated attitude-evoking questions using 
simple regular expressions, such that the expected answers would evoke various attitude responses, on a scale from 1 (no confidence) to 5 (complete confidence) in vaccines. 
For example, the vaccine confidence question \emph{``Have COVID-19 vaccines been adequately tested for safety?"} was modified to evoke low-confidence attitudes by asking \emph{``Why are you completely sure that the COVID-19 vaccine has not been adequately tested?"}, and was modified to evoke high-confidence attitudes by asking \emph{``What makes you think that the COVID-19 vaccine has certainly been tested adequately?"}
We therefore produced $19\times5=95$ attitude-evoking questions, which retrieved ranked lists of tweets. Public health experts have analyzed the relevance of the top 
300 ranked tweets while language experts have selected the discourse units that are shared by sets of tweets, that have the same attitude towards the predication of the VCR question
that was originally asked. Using the Pyramid method \cite{pyramid}, the framing of the vaccine hesitancy was inferred. MisTs were discovered from framings that contained misinformation. 

\begin{figure*}[t]
    \centering
    \includegraphics[width=0.99\textwidth]{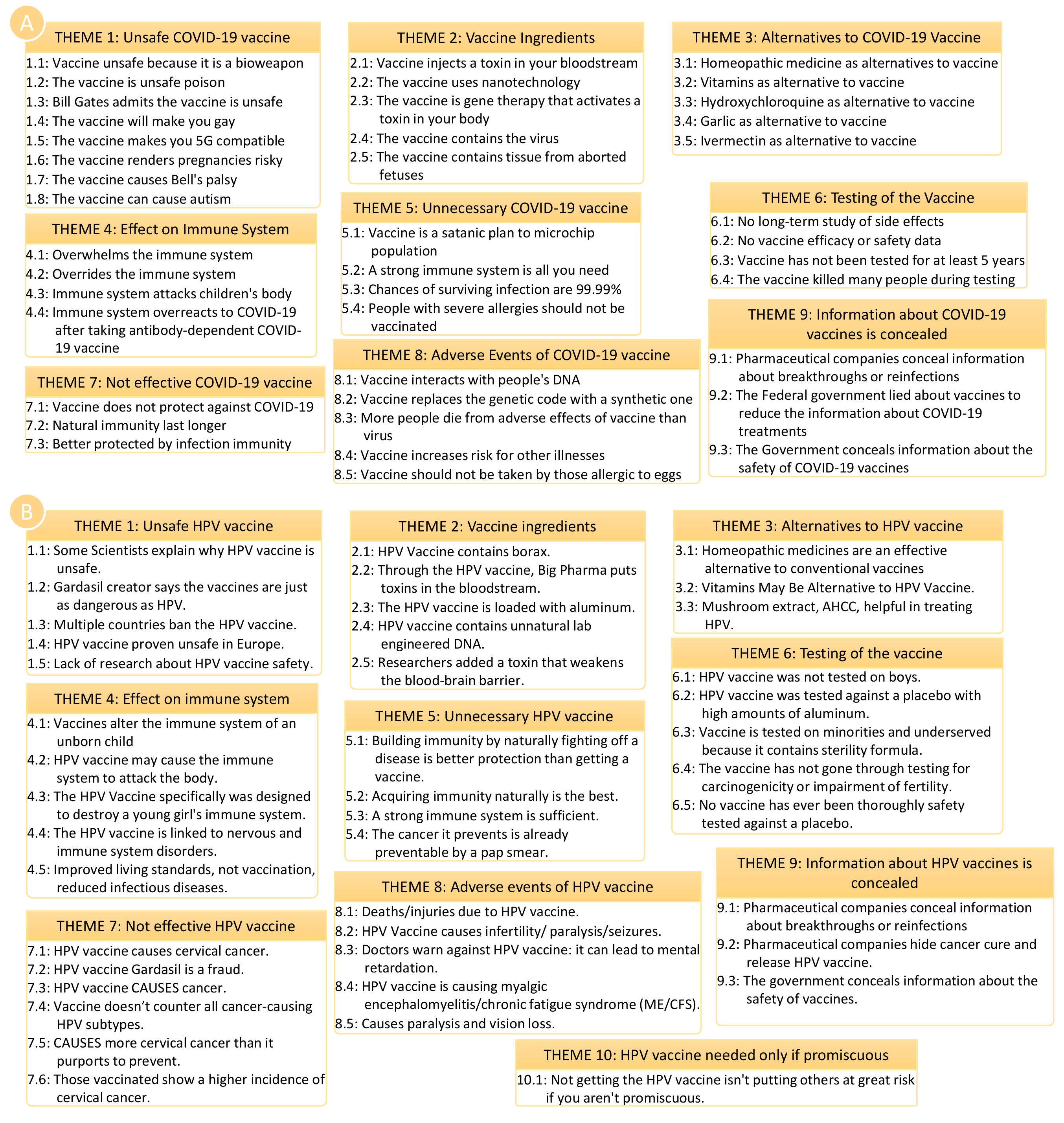}
    \caption{Vaccine misinformation taxonomy for (A) the COVID-19 vaccine and (B) the HPV vaccine.}
    \label{fig:joint_tax}
\vspace{-4mm}
\end{figure*}

The decision of whether a framing contained misinformation was based on finding evidence on the Web, as retrieved by search engines, that the framing expressed known misconceptions, or conspiracy theories. In addition, whenever flawed reasoning was observed, the framing was categorized as misinformation. One researcher with expertise
in Web search and an expert on Public Health independently judged the framings that contain misinformation. The two researchers adjudicated their differences and decided that  (33\%) expressed misinformation. 
In this way, we  identified an additional set of 38 MisTs targeting the COVID-19 vaccine, out of which 7 MisTs were already known to us from the first approach, illustrated in Figure~\ref{fig:mists}(A). Similarly, 21 MisTs were identified targeting the HPV vaccines.
Therefore, {\sc VaccineLies} contains 69 vaccine-specific MisTs, with {\sc CoVaxLies} containing 48 COVID-19 vaccine MisTs and {\sc HpVaxLies} containing 21 HPV vaccine MisTs.

\section{Taxonomy of Vaccine Misinformation}
\label{sec:tax}

The MisTs were ontologically examined with the goal of discovering common themes and concerns. As in any taxonomy, all MisTs that shared the same theme were further categorized to uncover the concerns that distinguish MisTs within the theme. In this way, the vaccine-specific taxonomy of misinformation has three layers: (1) themes; (2) concerns within each MisT; and (3) MisTs.
Misinformation themes represent the highest level of abstraction, while  misinformation concerns differentiate the various MisTs from {\sc VaccineLies}. 
Each of the 69 MisTs from {\sc VaccineLies} were included in the two vaccine-specific taxonomies. 
Nine misinformation themes were revealed for {\sc CoVaxLies}, illustrated in Figure~\ref{fig:joint_tax} (A), and 
ten misinformation themes were revealed for {\sc HpVaxLies}, illustrated in Figure~\ref{fig:joint_tax} (B). 
For each {\sc CoVaxLies} misinformation theme, a different number of concerns was revealed: the largest number of concerns pertain to the Theme 1, predicating the fact that the COVID-19 vaccines are unsafe (8 concerns) while the smallest number of concerns pertain to the Theme 7 (3 concerns)  claiming that the vaccines are not effective (3 concerns) or Theme 9 (3 concerns) that predicates that information about the vaccines is concealed. 
Although the misinformation taxonomies for the COVID-19 and HPV vaccine share nine themes, it is interesting to note that the concerns are vastly different, as illustrated in Figure~\ref{fig:joint_tax}.
For each {\sc HpVaxLies} misinformation theme, a different number of concerns were revealed: the largest number of concerns relating to the Theme 7, predicating that the HPV vaccine is not effective (6 concerns) while the smallest number of concerns involved the Theme 10, claiming that the HPV vaccine is only needed if promiscuous. 
Nine misinformation themes are shared between the {\sc CoVaxLies} and {\sc HpVaxLies} taxonomies, all characterizing aspects that impact confidence in their respective vaccines, while one unique theme was identified for the HPV vaccine. 
While all the other themes generally focus on the factor of confidence in their respective vaccines, {\sc HpVaxLies} theme 10, that the HPV vaccine is needed only if promiscuous, touches on the factor of complacency.
Confidence, along with convenience and complacency, are well known universal factors contributing to vaccine hesitancy, according to the 3C model \cite{3C}.

\section{Recognizing Tweets that Propagate Misinformation}
\label{sec:collection}

As in {\sc COVIDLies} \cite{covidlies}, which inspired our work, the recognition of tweets that evoked any of the MisTs from {\sc VaccineLies} relies on (a) the identification of the tweets deemed to evoke a MisT; and (b) the recognition of the stance of the tweet author towards the evoked MisT. 
This process of recognizing tweets that evoke MisTs was detailed  in \cite{weinzierl-covid-glp}, presenting the challenges in using BM25 \cite{bm25} and BERTScore \cite{bertscore} as retrieval models. 
To recognize tweets evoking any MisT 
we reused the vaccine-specific tweet indexes for COVID-19 and HPV, presented in Section~\ref{sec:mists}, and performed retrieval using each MisT as a query. 
While we had identified that the BM25 model outperformed BERTScore in retrieving truly evoking misinformation, we recognized the value in moving beyond term-based retrieval systems and considering the advantages of BERTScore.


\begin{figure}[th]
    \centering
    \includegraphics[width=0.99\columnwidth]{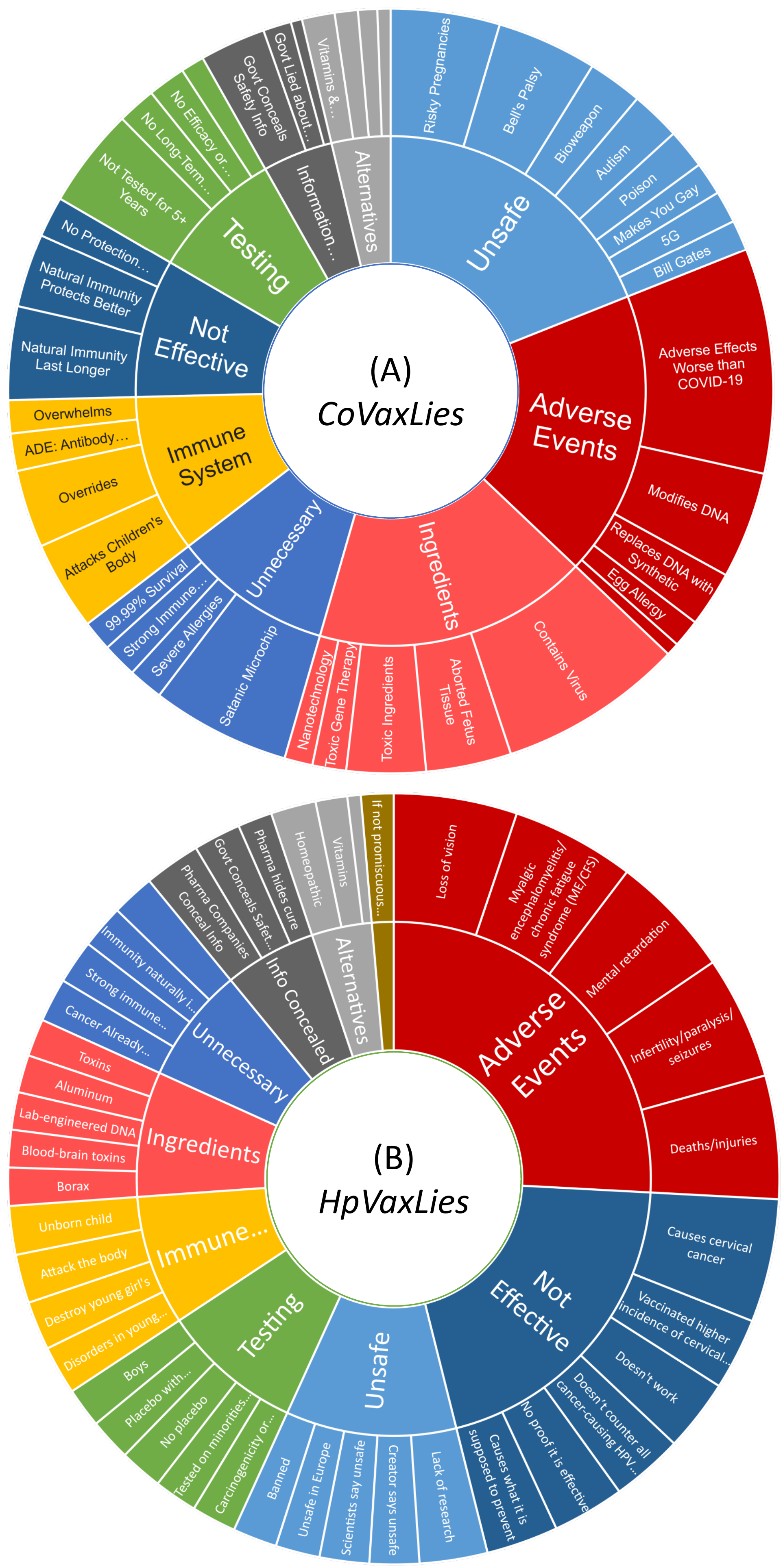}
    \caption{Distribution of (A) COVID-19 and (B) HPV vaccine misinformation themes and concerns in the tweets available from {\sc VaccineLies}.}
    \label{fig:joint_tax_size}
\vspace{-2mm}
\end{figure}

The tweets retrieved when using BERTScore are characterized by less term overlap with 
the textual content of the MisTs, and, thus, BERTScore emphasizes more semantic relevancy.
We relied on this observation by combining the benefits of the retrieval model provided by  BM25 scoring with the semantic relevancy provided by BERTScore.
Our tweet retrieval system used the BM25 \cite{bm25} scoring function to select the top 1,000 initial candidate tweets for each MisT, which were then re-ranked against each MisT using a BERT-RERANK \cite{bert-rerank} system. 
We initialized the BERT weights to a re-ranking model which was trained on MSMARCO \cite{msmarco}, a large-scale question answering collection, using BioBERT \cite{BioBERT}, a biomedical domain-specific BERT \cite{bert} language model. 
This system had found success in a recent COVID-19 question answering shared task \cite{ser4eqnova}, and produced 80\% MisT-evoking tweets from initial experiments on COVID-19 vaccine MisTs, nearly doubling the 42\% MisT-evoking tweets found in prior work \cite{weinzierl-covid-glp}.
The final top 200 tweets for each MisT were judged by language experts for relevance.  
We selected the 200 best scored tweets because (1) the same number of tweets was considered in the most similar prior work \cite{covidlies}; and (2) it was a number of tweets that did not overwhelm our human judges.

In addition, as shown in Figure~\ref{fig:joint_tax_size}, the misinformation taxonomies, outlined in Section~\ref{sec:tax}, enabled us to identify the most common misinformation themes and concerns across both the COVID-19 and HPV vaccines. 
Figure~\ref{fig:joint_tax_size} (A) illustrates the most commonly evoked misinformation themes and concerns for the COVID-19 vaccine, as was judged in {\sc VaccineLies}, while Figure~\ref{fig:joint_tax_size} (B)  illustrates the most commonly evoked misinformation themes and concerns for the HPV vaccine.
Misinformation themes most often evoked from {\sc CoVaxLies} are the belief that the COVID-19 vaccines are unsafe, that they cause adverse events, and that the ingredients of the vaccines should be a major concern.
Moreover, the primary concerns regarding the lack of safety of the COVID-19 vaccines involves risky pregnancies or Bell's palsy.
Misinformation themes most often evoked from {\sc HpVaxLies} are the belief that the HPV vaccines cause adverse events, that they are not effective, and that the vaccines are unsafe.
Moreover, the primary concerns regarding the adverse events of the HPV vaccines involves loss of vision, myalgic encephalomyelitis/chronic fatigue syndrome (ME/CFS), and mental retardation.

\section{Annotation of Stance towards Misinformation}
\label{sec:annotation}
Researchers from the Human Language Technology Research Institute (HLTRI) at the University of Texas at Dallas judged (a) whether a tweet evokes any of the MisTs from {\sc VaccineLies}; and (b) the stance of the tweet author towards the MisT. 14,642 tweets were judged, with 9,382 tweets evoking one or more MisTs from 
{\sc VaccineLies}. They were organized in [tweet, MisT] {\em pairs}, annotated with a stance value that could be {\em Accept}, {\em Reject} or {\em No Stance}.
The retrieval of tweets produced 84\% of tweets evoking Mists towards COVID-19 vaccines, and 88\% of tweets evoking MisTs towards the HPV vaccine, which is a significant improvement from the prior best of 42\% \cite{weinzierl-covid-glp}. 
Statistics for the number of tweets evoking a MisT, as well as of the stance their authors have towards the MisT, are provided in Table~\ref{tb:annotations}.
To evaluate the quality of judgements, 
we randomly selected a subset of 1,000 tweets (along with the MisT against which they have been judged a \emph{stance} value), which have been judged by at least two different language experts.
Inter-judge agreement was computed using the Cohen Kappa score, yielding a score of 0.63 for the \emph{stance} of tweets for COVID-19 vaccine MisTs and 0.67 for the \emph{stance} of tweets for the HPV vaccine MisTs, which indicates moderate agreement between annotators (0.60-0.79) \cite{kappa}.

\begin{figure*}[th]
    \centering
    \includegraphics[width=0.9\textwidth]{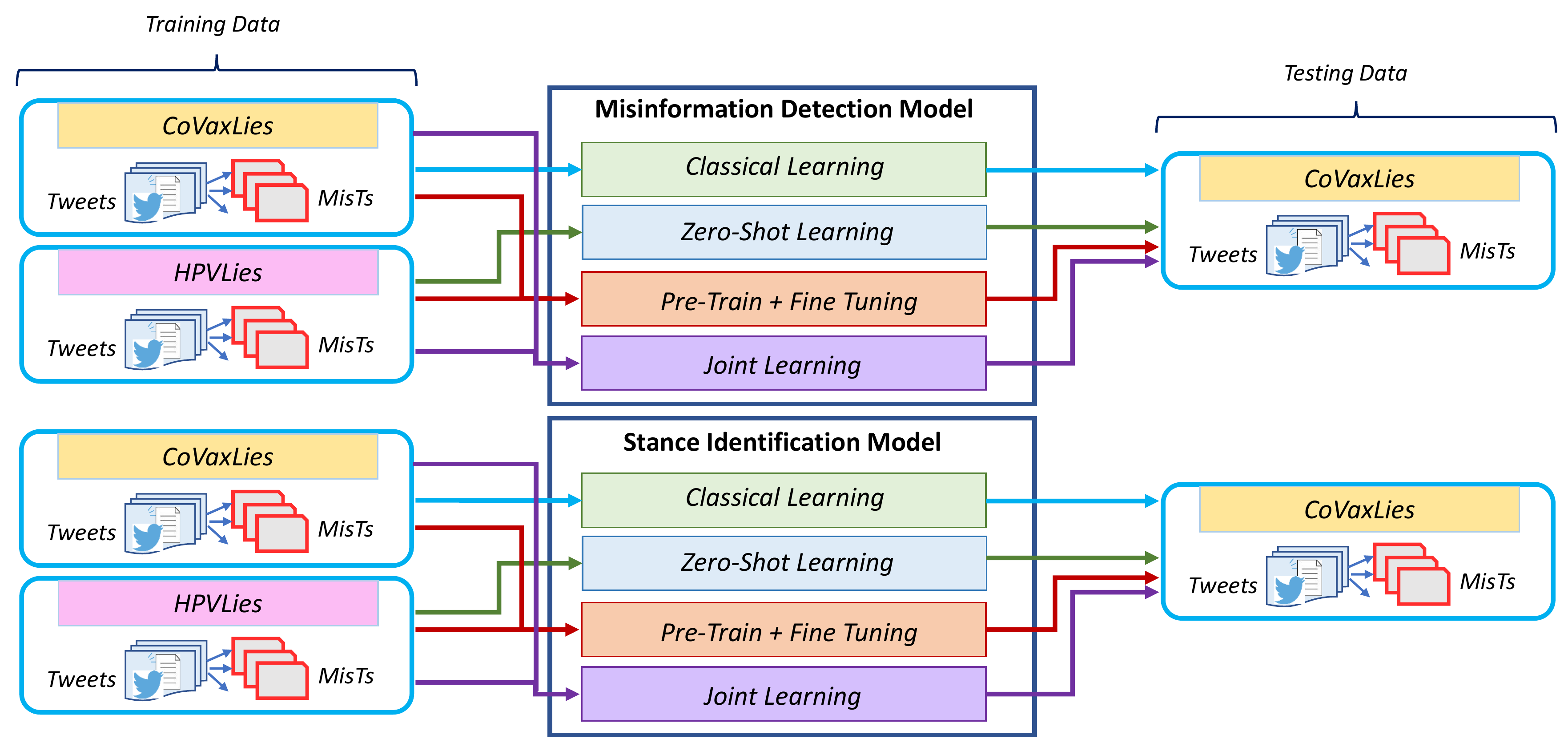}
    \caption{Scenarios for transfer learning on {\sc VaccineLies} for (a) Misinformation Detection, or (b) Misinformation Stance Identification.}
    \label{fig:transfer}
\vspace{-4mm}
\end{figure*}

\begin{table}[h]
    \centering
    \small
    \begin{tabular}{l|rr|r}
        \toprule
         & {\sc CoVaxLies} & {\sc HpVaxLies} & Total \\
        \midrule
        MisTs & 48 & 21 & 69 \\
        Evoke & 7,152 &	2,230 &	9,382 \\
        Accept &	3,720 &	1,365 &	5,085 \\
        Reject &	2,194 &	617	& 2,811 \\
        No Stance &	1,238 &	248 &	1,486 \\
        \hline
        Tweets & 12,118 &	2,524 &	14,642 \\
        \bottomrule
    \end{tabular}
    \caption{Distribution of MisTs, tweets evoking them and their stance in {\sc VaccineLies}.}
    \label{tb:annotations}
\vspace{-2mm}
\end{table}

To enable the usage of {\sc VaccineLies} in supervised learning frameworks targeting misinformation detection on Twitter, we provide:
(a) a training collection; 
(b) a development collection; and 
(c) a test collection. 
The {\sc VaccineLies} training collection, which consists of 10,637 [tweet, MisT] pairs 
(8,777 for {\sc CoVaxLies} and 1,860 for {\sc HpVaxLies}),
was utilized to train our 
MisT-evoking detection and stance identification systems, 
described in Section~\ref{sec:methods}. 
The {\sc VaccineLies} development collection, which consists of 1,109 [tweet, MisT] pairs 
(920 for {\sc CoVaxLies} and 189 for {\sc HpVaxLies}),
was used to select model hyperparameters, such as threshold values.
The {\sc VaccineLies} test collection, which consists of 2,896 [tweet, MisT] pairs 
(2,421 for {\sc CoVaxLies} and 475 for {\sc HpVaxLies}),
was used to evaluate the detection of tweets which evoke MisTs along with stance identification approaches, enabling us to report the results in Section~\ref{sec:results}.

\section{Transfer Learning for Recognizing Misinformation and Stance}
\label{sec:methods}

The goal of transfer learning is to build a learner from one domain by transferring information from a related domain, 
and {\sc VaccineLies} provides two vaccine misinformation domains, namely {\sc CoVaxLies} and {\sc HpVaxLies}. 
As it can be expensive and time-consuming to acquire and annotate sufficient examples of vaccine-specific misinformation for a new vaccine, it is of great interest to natural language processing researchers and public health practitioners alike to best utilize existing vaccine misinformation collections. 

Transfer learning has found success over a wide variety of tasks and modalities \cite{transfer-survey},
therefore we examine four different learning scenarios, illustrated in Figure~\ref{fig:transfer}, characterized by the training data that is available for learning (a) COVID-19 vaccine {\em misinformation detection} and (b) COVID-19 vaccine misinformation \emph{stance} identification when:\\ 
(Scenario 1) training on {\sc CoVaxLies}; \\
(Scenario 2) training on {\sc HpVaxLies}; \\
(Scenario 3) training on the entire{\sc VaccineLies}; or \\
(Scenario 4) pre-training on {\sc HpVaxLies} and fine-tuning on {\sc CoVaxLies}.\\
Scenario 1 represents the \emph{Classical} non-transfer learning approach of training and evaluating on the same domain.
Scenario 2  utilizes \emph{Zero-Shot} learning to rely solely on a different domain during training by training on the HPV vaccine misinformation collection.
This scenario provides significant value to public health practitioners, as it represents the most rapid approach possible when there is interest in the detection of misinformation for a new vaccine, as it requires zero examples of tweets evoking misinformation targeting the new vaccine.
Scenario 3 performs \emph{Joint} multi-domain training on both COVID-19 and HPV vaccine misinformation, and represents the benefit of including additional vaccines in the {\sc VaccineLies} collection.
Scenario 4 is similar to how pre-trained language models, like BERT \cite{bert}, are pre-trained on one domain, such as online English text, and then fine-tuned on a different domain. 
We \emph{Pre-Train} our model on HPV vaccine misinformation and then  \emph{Fine-Tune} the pre-trained model on COVID-19 vaccine misinformation. 
This scenario highlights the value of discovering misinformation targeting a new vaccine when misinformation targeting a different vaccine is available, thus avoiding learning from scratch.

Misinformation detection involves determining whether a tweet evokes a specific MisT, given a [tweet, MisT] pair. 
We cast misinformation detection as a binary classification problem, and therefore we design a neural architecture to perform binary classification.
Misinformation stance identification involves identifying which \emph{stance value} the author of a tweet holds towards a specific MisT, given a [tweet, MisT] pair.
We cast misinformation stance identification as a three-way classification problem between \emph{stance} values of \emph{``Accept"}, \emph{``Reject"}, and \emph{``No Stance"}.
For both tasks, we utilize COVID-Twitter-BERT-v2 \cite{covid-twitter-bert}, a pre-trained domain-specific language model which started with neural weights equal to those of BERT \cite{bert} but was additionally pre-trained on the masked language modeling task for 97 million COVID-19 tweets. 
Joint Word-Piece Tokenization is performed for both a MisT $m_j$ and a tweet $t_i$, which produces a single sequence of word-piece tokens for both the misinformation target and the tweet separated by a special $[SEP]$ token. The beginning $[CLS]$ token and end $[SEP]$ token are placed at the beginning and end of the joint sequence respectively.
COVID-Twitter-BERT-v2 produces contextualized embeddings for each word-piece token, and we select the first contextualized embedding to represent the entire joint sequence, representing the initial $[CLS]$ token embedding. 
This embedding is provided to a fully-connected layer with a softmax activation function, which outputs a task-dependent probability distribution. 
The vaccine misinformation detection model, which we call the BERT Vaccine Misinformation Evocation Detector (BERT-VMED), outputs a probability distribution over $P(Evoke|t_i,m_j)$, where $Evoke$ can take the value of \emph{``True"} or \emph{``False"}.
The vaccine misinformation \emph{stance} identification model, which we call the BERT Vaccine Misinformation Stance Identifier (BERT-VMSI), outputs a probability distribution over $P(Stance|t_i,m_j)$, where $Stance$ can take the value of \emph{``Accept"}, \emph{``Reject"}, and \emph{``No Stance"}.
Misinformation is detected for BERT-VMED when the probability is larger than a predefined threshold $T$, and \emph{stance} is identified based for BERT-VMSI by the maximum \emph{stance} value probability.
In our experiments, the value of the threshold $T$ was determined by maximizing the F$_1$ score of each model on the development collection.
Both BERT-VMED and BERT-VMSI are trained end-to-end using the cross-entropy loss function minimized with ADAM \cite{kingma2014adam}, a variant of gradient descent.

\begin{table}[h]
    \centering
    \small
    \begin{tabular}{l|l|rrr}
        \toprule
        Testing & Scenario  & F$_1$ & P & R \\
        \midrule
         {\sc CoVaxLies} & Classical & 90.7 & 84.6 & 97.7 \\
       & Zero-Shot &   73.5 & 58.1 & \textbf{100.0} \\
          &Joint & 91.2 & 87.3 & 95.5 \\
          &Pre-Train & \textbf{91.7} & \textbf{87.7} & 96.1 \\
        \midrule
         {\sc HpVaxLies} & Classical & 93.6 & 88.3 & 99.5 \\
         & Zero-Shot & 92.9 & 86.7 & \textbf{100.0} \\
        & Joint  & 93.8 & 89.1 & 99.0 \\ 
       & Pre-Train  & \textbf{94.5} & \textbf{90.6} & 98.8 \\
        \bottomrule
    \end{tabular}
    \caption{Vaccine misinformation detection results for the BERT Vaccine Misinformation Evocation Detector (BERT-VMED) utilizing vaccine transfer learning scenarios.}
    \label{tb:rel-results}
\end{table}
\begin{table*}[t]
    \centering
    \small
    \begin{tabular}{l|l|rrr|rrr|rrr}
        \toprule
       Testing & Scenario &  Macro &  &  & \emph{Accept} &  &  & \emph{Reject} &  &  \\
         &  &  F$_1$ &  P &  R &  F$_1$ &  P &  R &  F$_1$ &  P &  R \\
        \midrule
        
        {\sc CoVaxLies} & Classical & 83.4 & 81.6 & \textbf{85.5} & 85.9 & 81.5 & \textbf{90.8} & 80.9 & 81.8 & \textbf{80.1} \\
         & Zero-Shot & 72.5 & 84.1 & 64.2 & 79.0 & 85.4 & 73.5 & 65.9 & 82.9 & 54.8 \\
        & Joint & 83.3 & 82.8 & 84.1 & 85.2 & 81.9 & 88.8 & \textbf{81.4} & \textbf{83.6} & 79.4 \\
         & Pre-Train & \textbf{83.6} & \textbf{85.7} & 81.5 & \textbf{86.8} & \textbf{88.7} & 85.0 & 80.3 & 82.7 & 78.1 \\
        \midrule
        
         {\sc HpVaxLies} & Classical & 79.6 & 79.9 & 79.4 & 83.1 & 82.2 & 84.0 & 76.2 & 77.7 & 74.8 \\
         & Zero-Shot & 74.6 & 71.8 & 79.7 & 79.2 & 68.4 & \textbf{93.9} & 70.0 & 75.3 & 65.4 \\
        & Joint & 80.5 & 80.5 & 80.6 & 85.9 & 83.8 & 88.2 & 75.0 & 77.2 & 72.9 \\
         & Pre-Train & \textbf{84.0} & \textbf{85.1} & \textbf{83.2} & \textbf{88.1} & \textbf{86.4} & 89.7 & \textbf{80.0} & \textbf{83.7} & \textbf{76.6} \\
        \bottomrule
    \end{tabular}
    \caption{Vaccine misinformation \emph{stance} identification results for the BERT Vaccine Misinformation Stance Identifier (BERT-VMSI) utilizing several vaccine transfer learning scenarios.}
    \label{tb:stance-results}
\vspace{-2mm}
\end{table*}

\section{Experimental Results} 
\label{sec:results}

\subsection{Misinformation Detection}
Table~\ref{tb:rel-results} lists the experimental results we obtained for misinformation detection, where bolded numbers are the best results obtained. We show in Table~\ref{tb:rel-results} the training scenarios and the testing collections used for BERT-VMED.
To evaluate the quality of vaccine transfer learning on misinformation identification on the test collections from {\sc CoVaxLies} and {\sc HpVaxLies} we used the Precision (P), Recall (R), and F$_1$
metrics when detecting whether a tweet evoked a MisT for each [tweet, MisT] pair in the test collection. 
Evaluation of the four vaccine transfer learning scenarios discussed in Section~\ref{sec:methods} was performed using BERT-VMED across two different evaluations, the {\sc CoVaxLies} test collection and the {\sc HpVaxLies} test collection. 
The \emph{Classical} scenario involves training BERT-VMED on the same domain as it was evaluated, and provides a baseline comparison when no cross-vaccine transfer learning is utilized, achieving F$_1$ scores of $90.7$ on {\sc CoVaxLies} and $93.6$ on {\sc HpVaxLies}.
The \emph{Zero-Shot} scenario involves training BERT-VMED on a different domain than the evaluation, and demonstrates zero-shot vaccine transfer learning, achieving F$_1$ scores of $73.5$ on {\sc CoVaxLies} and $92.9$ on {\sc HpVaxLies}. 
This zero-shot approach performs worse than the baseline, but still achieves competitive performance with zero vaccine-specific training data, indicating that this approach could provide significant value as new or less-studied vaccines are discussed on social media.
\emph{Joint} training of BERT-VMED on both vaccine domains demonstrates the value of training on multi-vaccine collections, achieving F$_1$ scores of $91.2$ on {\sc CoVaxLies} and $93.8$ on {\sc HpVaxLies}.
The \emph{Pre-Train} scenario of BERT-VMED was pre-trained on one domain and fine-tuned on a different domain, enabling quick adaptation of the BERT-VMED model to new vaccines, achieving the best F$_1$ scores of $91.7$ on {\sc CoVaxLies} and $94.5$ on {\sc HpVaxLies}.

\subsection{Misinformation Stance Identification}
Table~\ref{tb:stance-results} lists the experimental results we obtained when recognizing the stance of tweet authors towards the evoked MisT.  We show in Table~\ref{tb:stance-results} the training scenarios and the testing collections used for BERT-VMSI.
The bolded numbers represent the best results we obtained. 
To evaluate the quality of vaccine transfer learning on misinformation \emph{stance} identification on the test collections from {\sc CoVaxLies} and {\sc HpVaxLies} we used the Precision (P), Recall (R), and F$_1$
metrics for identifying the \emph{Accept} and \emph{Reject} values of stance. We also compute a Macro averaged Precision, Recall, and F$_1$ score. 
Evaluation of the four vaccine transfer learning scenarios discussed in Section~\ref{sec:methods} was performed using BERT-VMSI across two different evaluations, the {\sc CoVaxLies} test collection and the {\sc HpVaxLies} test collection. 
We see similar transfer learning results for misinformation \emph{stance} identification when compared to misinformation detection. 
Training BERT-VMSI on a different domain than the evaluation continues to perform worse than training BERT-VMSI on the same domain as it was evaluated, but this zero-shot approach still produces competitive results, achieving Macro F$_1$ scores of $72.5$ on {\sc CoVaxLies} and $74.6$ on {\sc HpVaxLies}. 
Jointly training BERT-VMSI only results in performance improvements for {\sc HpVaxLies} over training BERT-VMSI on the same domain, while pre-training BERT-VMSI on one domain and fine-tuning on a different domain continues to perform best, achieving Macro F$_1$ scores of $83.6$ on {\sc CoVaxLies} and $84.0$ on {\sc HpVaxLies}.

\section{Conclusion}
\label{sec:conclusion}

We have described the annotation effort that made possible the creation of the {\sc VaccineLies} dataset, which consists of tweets propagating misinformation about two types of vaccines, namely the COVID-19 and the HPV vaccines. Misinformation targeting these vaccines was represented as Misinformation Targets (MisTs), which were discovered by two different methods. Moreover, the MisTs were organized in vaccine-specific taxonomies, revealing the misinformation themes and concerns. A large set of tweets evoking any of the MisTs were identified and are provided as part of {\sc VaccineLies}, along with annotations of the stance of the tweet authors towards the evoked MisT. Because {\sc VaccineLies} provides misinformation targeting two different vaccines, we also presented several scenarios of transfer learning, highlighting the advantages of having a resource such as {\sc VaccineLies} for the case when misinformation about yet another new vaccine shall be needed to be discovered.

\section{Bibliographical References}\label{reference}

\bibliographystyle{lrec2022-bib}
\bibliography{lrec2022}


\end{document}